\pdfoutput=1

\documentclass[11pt]{article}

\usepackage{naacl2021}

\usepackage{times}
\usepackage{latexsym}
\usepackage{graphicx}

\usepackage{array}
\usepackage{mathtools}
\usepackage{subcaption}
\usepackage{pbox}
\usepackage{makecell}
\usepackage{float}
\usepackage{tabularx}
\usepackage[group-separator={,},group-minimum-digits={3}]{siunitx}
\newcolumntype{P}[1]{>{\centering\arraybackslash}p{#1}}
\usepackage{scalerel,xparse}
\NewDocumentCommand\emojismile{}{
    \includegraphics[scale=0.020]{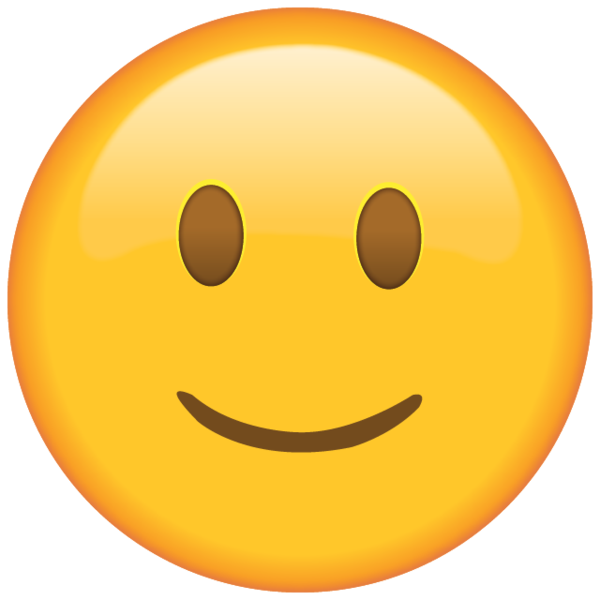}
}

\NewDocumentCommand\emojisad{}{
    \includegraphics[scale=0.020]{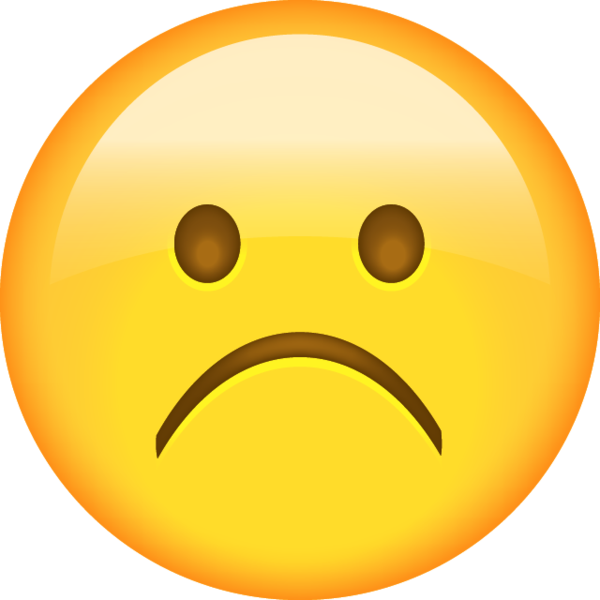}
}

\usepackage[T1]{fontenc}

\usepackage[utf8]{inputenc}

\usepackage{microtype}

\title{CaSiNo: A Corpus of Campsite Negotiation Dialogues for Automatic Negotiation Systems}

\author{Kushal Chawla$^{1}$\hspace{0.3cm}Jaysa Ramirez$^{2}$\thanks{\hspace{0.5mm} Work done when authors were interns at USC ICT} \hspace{0.1mm}\hspace{0.3cm} Rene Clever$^{3*}$\hspace{0.3cm} Gale Lucas$^1$\hspace{0.3cm}\\ \textbf{Jonathan May}$^4$\hspace{0.3cm} \textbf{Jonathan Gratch}$^1$ \\
$^{1 \& 4}$University of Southern California, Los Angeles, USA \\
$^2$Rollins College, Winter Park, USA \hspace{0.3cm}$^3$CUNY Lehman College, Bronx, USA  \\
$^1$\texttt{\{chawla,lucas,gratch\}@ict.usc.edu}\\ $^2$\texttt{jramirez@rollins.edu}\hspace{0.3cm}  $^3$\texttt{rene.clever@lc.cuny.edu}\hspace{0.3cm}
\\$^4$\texttt{jonmay@isi.edu}}

\begin{document}
\maketitle
\begin{abstract}
Automated systems that negotiate with humans have broad applications in pedagogy and conversational AI. To advance the development of practical negotiation systems, we present CaSiNo: a novel corpus of over a thousand negotiation dialogues in English. Participants take the role of campsite neighbors and negotiate for food, water, and firewood packages for their upcoming trip. Our design results in diverse and linguistically rich negotiations while maintaining a tractable, closed-domain environment. Inspired by the literature in human-human negotiations, we annotate persuasion strategies and perform correlation analysis to understand how the dialogue behaviors are associated with the negotiation performance. We further propose and evaluate a multi-task framework to recognize these strategies in a given utterance. We find that multi-task learning substantially improves the performance for all strategy labels, especially for the ones that are the most skewed. We release the dataset, annotations, and the code to propel future work in human-machine negotiations: \url{https://github.com/kushalchawla/CaSiNo}.

\end{abstract}

\section{Introduction}
Negotiations are highly prevalent in our interactions, from deciding who performs the household chores to high-stake business deals to maintaining international peace. Automatic negotiation systems are helpful in providing cost-effective social skills training~\cite{johnson2019intelligent} and for advanced capabilities of AI assistants such as Google Duplex~\cite{leviathan2018google}.

A negotiation requires understanding the partner's motives along with effective reasoning and communication, which is challenging for an automated system. Prior work in human-machine negotiations primarily uses strict communication protocols such as a pre-defined menu of options~\cite{mell2016iago}. Systems involving free-form dialogue are limited due to a lack of interdisciplinary efforts in NLP and Computational Social Science in this direction. Initial efforts in building dialogue systems for negotiations looked at game environments~\cite{asher2016discourse, lewis2017deal}. DealOrNoDeal~\cite{lewis2017deal} involves two negotiators who split given quantities of three arbitrary items: books, balls, and hats. This provides a concrete structure to the negotiation, keeps the design tractable, and ensures a reliable evaluation based on final points scored. Many practical solutions in negotiations follow similar \textit{closed-domain} designs~\cite{mell2016iago}. However, most of the dialogues in these game settings reduce to merely an exchange of offers from both sides. For instance, `\textit{i need the book and the balls you can have the hat}' or `\textit{i want the ball and $2$ books}' in DealOrNoDeal. One reason for this lack of richness in language use is that the items are \textit{arbitrarily defined}, that is, there is \textit{no semantic context} around the items that the participants are negotiating for. Hence, this setup fails to capture many realistic aspects of negotiations such as small talk, preference elicitation, emotion expression, and convincing strategies based on individual preferences and requirements. Emulating real-world negotiations is desirable for developing practical systems for social skills training and robust AI assistants that are useful in realistic scenarios.

On the other extreme, the CB dataset~\cite{he2018decoupling} involves buyer-seller negotiations to finalize the price of a given product. Targeting the collection of more \textit{open-ended} dialogues, the participants are also encouraged to discuss side offers, such as free delivery or also selling other accessories at the same price. Although this promotes diversity and rich natural conversations, unfortunately, such open-ended domains make the evaluation of negotiation performance non-trivial, which also inhibits the practical applicability of the systems developed on such datasets. For instance, in skills training, it is desirable to judge the performance and provide critical feedback~\cite{monahan2018autonomous}.

To address these shortcomings, we design a novel negotiation task. Our design is based on a tractable \textit{closed-domain} abstraction from the negotiation literature but is infused with a real-world camping scenario, resulting in rich dialogues for natural language research (Section \ref{sec:corpusdesign}). The task involves two participants who take the role of campsite neighbors and negotiate for additional \textit{Food}, \textit{Water}, and \textit{Firewood}, based on individual preferences and requirements. 

Based on this design, we collect \textbf{CaSiNo}: a corpus of $1030$ \textbf{Ca}mp \textbf{Si}te \textbf{N}eg\textbf{o}tiation dialogues in English. The dialogues contain various aspects of a realistic negotiation, such as rapport building, discussing preferences, exchanging offers, emotion expression, and persuasion with personal and logical arguments. We also collect the participants' satisfaction from the outcome and how much they like their opponents, both being important metrics in negotiations~\cite{mell2019likeability}. We annotate $9$ persuasion strategies that span cooperative to selfish dialog behaviors (Section \ref{sec:data-anns}). We perform an extensive correlational analysis to investigate the relationship among the final outcomes and explore how they relate to the use of negotiation strategies (Section \ref{sec:analysis}). Further, we propose a multi-task framework with task-specific self-attention mechanisms to recognize these strategies in a given utterance (Section \ref{sec:modelling}). Our insights form the foundation for the development of practical negotiation systems that engage in free-form natural conversations. We release the dataset along with the annotations to enable future work in this direction. 

\section{The CaSiNo Dataset}
\label{sec:corpusdesign}

\begin{table*}[th]
\centering
\scalebox{0.8}{
\begin{tabular}{p{9cm}|p{9cm}}
\hline
\multicolumn{2}{c}{\textbf{Preferences \& Arguments}} \\ \hline
\multicolumn{1}{c|}{\textbf{P1}} & \multicolumn{1}{c}{\textbf{P2}}\\ \hline
\textit{High}: Water: We like to go on runs and it increases the need of this. & \textit{High}: Food: Food really increases everyones morale.\\ \hline
\textit{Medium}: Food: Food overall is a good mood booster. & \textit{Medium}: Firewood: We like to have a large fire.\\ \hline
\textit{Low}: Firewood: We do not care for fire and it is not necessary to us. & \textit{Low}: Water: We don't drink water that often.\\ \hline
\end{tabular}}\\
\vspace{0.3cm}
\scalebox{0.8}{
\begin{tabular}{p{15cm}|p{3cm}}
\multicolumn{1}{c|}{\textbf{Conversation}} & \multicolumn{1}{c}{\textbf{Annotation}} \\ \hline
\textbf{P1}: How are you today? Did you have any preferences on the supplies we will be trading? & Small-Talk, Coordination, Elicit-Pref\\ \hline
\textbf{P2}: I am good. How about yourself? I think I would like some firewood to start off with. We like to have bigger fires. What about you? & Small-Talk, Self-Need, Other-Need, Elicit-Pref\\ \hline
\textbf{P1}: I am good as well. That is good to hear that you like to have bigger fires as we do not care much for that. We would much rather have some extra water. & Small-Talk, Empathy, No-Need\\ \hline
\textbf{P2}: Water is a little important to us too though , if possible maybe we can split that or maybe we can get some more food in replacement. & Coordination\\ \hline
\textbf{P1}: That may be possible....\emojismile What did you have in mind for the food replacement? & Non-strategic\\ \hline
\textbf{P2}: You can have all the water if we can have all the food? & Non-strategic\\ \hline
\textbf{P1}: I dont think I am okay with that \emojisad. Food is essential to our groups morale when camping. We would like 1 additional food preferably.& Self-Need, Other-Need\\ \hline
\textbf{P2}: Well you guys did say you did not care much about large fires. What if you gave all the firewood in replace for the water and you can still keep 1 food? & UV-Part, Coordination\\ \hline
\textbf{P1}: So I would get 3 water and 1 food and youd get 3 firewood and 2 food?& Non-strategic\\ \hline
\textbf{P2}: Yea that seems like an alright trade to me \emojismile& Non-strategic\\ \hline
\textbf{P1}: Hmm... alright then \emojismile & Non-strategic\\ \hline
\textbf{P2}: Submit-Deal & \\ \hline
\textbf{P1}: Accept-Deal & \\ \hline
\end{tabular}}
\caption{Sample dialogue from the CaSiNo dataset. \textbf{P1} and \textbf{P2} represent two participants in our study.}
\label{tab:sample-dialogue}
\end{table*}

Our data was crowd-sourced on Amazon Mechanical Turk. We describe our design by following the journey of a specific participant in our study.

\noindent \textbf{Pre-Survey}: We start by collecting demographics and psychological personality traits of the participants which relate to their negotiation behaviors. For demographics, we gather age, gender, ethnicity, and the highest level of education. We consider two measures of individual personality differences: Social Value Orientation or SVO~\cite{van1997development} and Big-5 personality~\cite{goldberg1990alternative} that have been heavily studied in the context of negotiations~\cite{bogaert2008social, curtis2015relationship}. SVO classifies the participants as \textit{Prosocial}, who tend to approach negotiations cooperatively, or \textit{Proself}, who tend to be more individualistic. Big-$5$ personality test assesses the participants on five dimensions: Extraversion, Agreeableness, Conscientiousness, Emotional Stability, and Openness to Experiences. Our participants exhibit diverse demography and psychological personality. We provide aggregate statistics in Appendix A.

\noindent\textbf{Negotiation Training}: 
Research shows that the average human is bad at negotiating~\cite{wunderle2007negotiate, babcock2009women}, which can adversely impact the quality of the collected dialogues and consequently, the system trained on them. One way to mitigate this is by using reinforcement learning to optimize on a reward that measures the negotiation performance. RL training has proved to be challenging and often leads to degeneracy~\cite{lewis2017deal}. Further, this ignores prior work in human-human negotiations that provides guidelines for achieving favorable outcomes in realistic negotiations~\cite{lewicki2016essentials}. 

To incorporate these best practices in a principled way, we design a training module. Each participant is asked to watch a video tutorial before their negotiation. The tutorial takes an example of a negotiation between two art collectors to encourage them to follow some of the best practices in negotiations~\cite{lewicki2016essentials}, including 1) Starting with high offers, 2) Discussing preferences, 3) Appropriate emotion expression, and 4) Discussing individual requirements to make convincing arguments. This results in a rich and diverse set of dialogues, as we explore further in later sections. We release the complete video tutorial publicly, with the hope that it promotes reproducibility and helps researchers to design similar data collection experiments in the future: \url{https://youtu.be/7WLy8qjjMTY}.

\noindent\textbf{Preparation Phase}: Several requirements guide our design choices: 1) \textit{Semantically Meaningful}: The context must be meaningful and relatable for MTurk participants and for anyone who negotiates with the system trained on this dataset. This allows the participants to indulge in personal and contextual conversations, making the resulting system more useful for downstream applications. 2) \textit{Symmetric task}: The task should be symmetric for both the participants so that a dialogue system may leverage both sides of the conversations during modelling, and 3) \textit{Symmetric items}: The items which the participants are negotiating for should be symmetric in the sense that an individual can resonate with any preference order assigned to them. Hence, every category of items can be more desirable over others depending on a real-world context.

Our scenario is an instance of a common and useful abstraction for studying negotiations in scientific literature known as the multi-issue bargaining task~\cite{fershtman1990importance}. The task involves campsite neighbors who negotiate for additional \textit{Food}, \textit{Water}, and \textit{Firewood} packages, each with a total quantity of three. Instead of choosing an arbitrary set of items, each item represents quite relatable, basic requirements that one might plausibly have for an actual camping trip. The items were only broadly defined to encourage diversity. One challenge when dealing with a realistic context like camping is the inherent bias that one might have towards one item over others, which violates our symmetry constraint. To mitigate this, we emphasize that the camping authorities have already provided the basic essentials and the participants will be negotiating for extras, based on their individual plans for camping. We present the negotiation scenario, as seen by participants, in Appendix B.

The three item types are assigned a random priority order for every participant using a permutation of \{\textit{High}, \textit{Medium}, \textit{Low}\}. As in realistic negotiations, the participants are asked to \textit{prepare} for their negotiation by coming up with justifications for the given preferences before the negotiation begins (precise question format in Appendix G), for instance, needing more water supplies for a hike or firewood for a bonfire with friends. We find that the participants are able to come up with a variety of arguments from their own camping experiences, such as \textit{Personal Care}, \textit{Recreational}, \textit{Group Needs} or \textit{Emergency} requirements. We illustrate some of these arguments in Appendix B. The participants were encouraged to use their justifications as they feel fit, to negotiate for a more favorable deal.

\noindent\textbf{Negotiation Dialogue}: Finally, two participants are randomly paired to engage in an alternating dialogue for a minimum total of $10$ utterances. We also provide the option to use emoticons for four basic emotions, namely, happy, sad, anger, and surprise. After coming to an agreement, the participants submit the deal formally using the provided options. They can also walk away from the negotiation if they are unable to come to an agreement. The primary evaluation metric to assess the negotiation performance is the number of points scored by a negotiator. Every \textit{High}, \textit{Medium}, and \textit{Low} priority item is worth $5$, $4$, and $3$ points respectively, such that a participant can earn a maximum of $36$ points if she is able to get \textit{all} the available items.

\noindent\textbf{Post-Survey}: We collect two other evaluation metrics relevant to negotiations: 1) $5$-point scale for satisfaction (How satisfied are you with the negotiation outcome?) and 2) $5$-point scale for opponent likeness (How much do you like your opponent?). Back-to-back negotiation~\cite{aydougan2020challenges} is an interesting case where the relationship with the partner is crucial. In such a case, a poor relationship in earlier negotiations can adversely impact the performance in later rounds. Further, for some cases in CaSiNo, we observed that the participants were satisfied with their performance, despite performing poorly because they thought that the arguments of their partners for claiming the items were justified. One might argue that this is still a successful negotiation. Hence, we believe that all the metrics defined in the paper are important in the context of real-world negotiations and propose that they should be looked at collectively. We will further analyze these outcome variables in Section \ref{sec:analysis} where we study the correlations between the participants' negotiation behaviors and these metrics of negotiation performance.

\noindent\textbf{Data Collection}: We collected the dataset over a month using the ParlAI framework~\cite{miller2017parlai}. Screenshots from the interface are provided in Appendix G. The participant pool was restricted to the United States, with a minimum $500$ assignments approved and at least $95\%$ approval rate. We post-process the data to address poor quality dialogues and inappropriate language use. We describe these post-processing steps in Appendix C. 

Finally, we end up with $1030$ negotiation dialogues between $846$ unique participants. On average, a dialogue consists of $11.6$ utterances with $22$ tokens per utterance. We present a sample dialogue with the associated participant profile in Table \ref{tab:sample-dialogue}. The participants are rewarded a base amount of $\$2$ for their time (around $20$ minutes). Further, they were incentivized with a performance-based bonus of $8.33$ cents for every point that they are able to negotiate for. If a participant walks away, both parties get the amount corresponding to one high item or the equivalent of $5$ points. The bonus is paid out immediately after the task to encourage participation. We discuss ethical considerations around our data collection procedure in Section \ref{sec:ethics}. Overall, the participants had highly positive feedback for our task and could relate well with the camping scenario, engaging in enjoyable, interesting, and rich personal conversations. We discuss their feedback with examples in Appendix D.


\section{Strategy Annotations}
\label{sec:data-anns}

\begin{table}[th]
\scalebox{0.7}{
\begin{tabular}{p{2.4cm}|p{5cm}|P{0.9cm}|c}
\hline
\textbf{Label} & \textbf{Example} & \textbf{Count} & \textbf{$\alpha$}\\ \hline
& \multicolumn{1}{c|}{\textbf{Prosocial Generic}} && \\
Small-Talk & Hello, how are you today? & $1054$ & $0.81$\\
Empathy & Oh I wouldn't want for you to freeze & $254$ & $0.42$\\
Coordination &  Let's try to make a deal that benefits us both! & $579$ & $0.42$\\ \hline
& \multicolumn{1}{c|}{\textbf{Prosocial About Preferences}} && \\
No-Need & We have plenty of water to spare. & $196$ & $0.77$\\
Elicit-Pref & What supplies do you prefer to take the most of? & $377$ & $0.77$\\ \hline
& \multicolumn{1}{c|}{\textbf{Proself Generic}} && \\
Undervalue-Partner & Do you have help carrying all that extra firewood? Could be heavy? & $131$ & $0.72$ \\
Vouch-Fairness & That would leave me with no water. & $439$ & $0.62$\\ \hline
& \multicolumn{1}{c|}{\textbf{Proself About Preferences}} && \\
Self-Need & I can't take cold and would badly need to have more firewood. & $964$ & $0.75$\\
Other-Need & we got kids on this trip, they need food too. & $409$ & $0.89$\\ \hline
Non-strategic & Hello, I need supplies for the trip! & $1455$ & - \\ \hline
\end{tabular}}
\caption{\label{tab:ann-stats} Utterance-level strategy annotations. $\alpha$ refers to Krippendorff's alpha among $3$ annotators on a subset of $10$ dialogues ($\sim120$ utterances). An utterance can have multiple labels.}
\end{table}

After collecting the dataset, we developed an annotation schema to analyze the negotiation strategies used by the participants, and to facilitate future work. We follow the conceptual content analysis procedure~\cite{krippendorff2004reliability} to design the scheme. Being a natural conversational dataset, we find several instances where a strategy spans multiple sentences in an utterance, as well as instances where the same sentence contains several strategies. Hence, we define an utterance as the \textit{level of analysis}. Each utterance is annotated with one or more labels. If no strategy is evident, the utterance is labelled as \textbf{Non-strategic}. Although we label entire utterances, self-attention shows some promise as an automatic way to identify which part of an utterance corresponds to a given strategy, if desirable for a downstream application (Section \ref{sec:modelling}).

Human negotiation behaviors can be broadly categorized as Prosocial, which promote the interests of others or the common good, and Proself, which tend to promote self-interest in the negotiations~\cite{yamagishi2017response,van2007self}. Another important criterion is discussing preferences. Prior work suggests that humans negotiate with a fixed-pie bias, assuming that the partner's preferences align, and hence achieving sub-optimal solutions~\cite{kelley1996classroom}. Based on these distinctions and manual inspection, we define $9$ strategies used in the CaSiNo dataset. The usage of these negotiation strategies correlates with both the objective and subjective metrics of negotiation performance.

\subsection{Prosocial}
Prosocial strategies address the concerns of both self and the negotiation partner. We define three strategies that exhibit \textit{generic Prosocial behavior}.

\textbf{Small-Talk}: Participants engage in small talk while discussing topics apart from the negotiation, in an attempt to build a rapport with the partner. For example, discussing how the partner is doing during the pandemic or sharing excitement for the camping trip. Rapport has been well studied to positively impact negotiation outcomes~\cite{nadler2003rapport}. Small talk usually appears either at the beginning or at the end of the negotiation.

\textbf{Empathy}: An utterance depicts \textbf{Empathy} when there is evidence of positive acknowledgments or empathetic behavior towards a personal context of the partner, for instance, towards a medical emergency. Empathy promotes Prosocial behaviors in interpersonal interactions~\cite{klimecki2019role}.

\textbf{Coordination} is used when a participant promotes coordination among the two partners. This can be, for instance, through an explicit offer of a trade or mutual concession, or via an implicit remark suggesting to work together towards a deal.

\noindent Further, we define two strategies that relate to \textit{Prosocial behavior about individual preferences}:

\textbf{No-Need} is when a participant points out that they do not need an item based on personal context such as suggesting that they have ample water to spare. \textbf{No-Need} can directly benefit the opponent since it implies that the item is up for grabs.

\textbf{Elicit-Pref} is an attempt to discover the preference order of the opponent. CaSiNo covers a range of scenarios based on how aligned the preferences of the two parties are. Generally, we find that discussing preferences upfront leads to smoother negotiations without much back and forth.

\subsection{Proself}
Proself behavior attempts to serve personal performance in a negotiation. We define two strategies exhibiting \textit{generic Proself behavior}.

\textbf{Undervalue-Partner} or \textbf{UV-Part}, refers to the scenario where a participant undermines the requirements of their opponent, for instance, suggesting that the partner would not need more firewood since they already have the basic supplies or a suggestion that there might be a store near the campsite where the partner can get the supplies instead.

\textbf{Vouch-Fairness} is a callout to fairness for personal benefit, either when acknowledging a fair deal or when the opponent offers a deal that benefits them. For instance, through an explicit callout `this deal is not fair', or implicitly saying `this does not leave me with anything'.

Finally, we consider two \textit{Proself strategies that relate to individual preferences}:  

\textbf{Self-Need} refers to arguments for creating a personal need for an item in the negotiation. For instance, a participant pointing out that they sweat a lot to show preference towards water packages.

\textbf{Other-Need} is similar to \textbf{Self-Need} but is used when the participants discuss a need for someone else rather than themselves. For instance, describing the need for firewood to keep the kids warm. Negotiating on behalf of others is densely studied as a competitive strategy, where negotiators engage in contentious, demanding, and inflexible bargaining behaviors~\cite{adams1976structure, clopton1984seller}.

\noindent\textbf{Collecting annotations}: Three expert annotators\footnote{Researchers involved in the project.} independently annotated $396$ dialogues containing $4615$ utterances. The annotation guidelines were iterated over a subset of $5$ dialogues, while the reliability scores were computed on a different subset of $10$ dialogues. We use the nominal form of Krippendorff's alpha~\cite{krippendorff2018content} to measure the inter-annotator agreement. We provide the annotation statistics in Table \ref{tab:ann-stats}. Although we release all the annotations, we skip \textbf{Coordination} and \textbf{Empathy} for our analysis in this work, due to higher subjectivity resulting in relatively lower reliability scores. For the rest of the paper, we will refer to this annotated subset of CaSiNo as CaSiNo-Ann.
\section{Correlational Analysis}
\label{sec:analysis}

We next perform correlational analysis on CaSiNo-Ann to understand how the points scored by a participant relate to their satisfaction from the outcome and their opponent perception. We further shed light on what kind of strategies are more likely to lead to better outcomes. Such insights motivate our experiments on strategy prediction and would direct future efforts in building negotiation systems. We present complete results in Appendix E and discuss the significant observations below.

\noindent\textbf{Relationship among outcome variables}: We consider the points scored, satisfaction from the outcome, and opponent likeness. We find that the points scored by a participant are positively correlated with their own satisfaction ($r$=$0.376$, $p<0.01$) and with their perception of the opponent ($r$=$0.276$, $p<0.01$). Similar trends are visible with the corresponding variables of the negotiation partner as well, suggesting that the participants secured more points while still maintaining a positive perception in the eyes of their opponents.

\noindent\textbf{Discovering the integrative potential}:
Integrative potential in a negotiation is based on how aligned the partner preferences are. Complete alignment leads to a \textit{distributive} (or zero-sum) negotiation, having a low integrative potential where the benefit of one results in a high loss for the other. A negotiation is \textit{integrative} if the preferences do not align, allowing for solutions that maximize mutual points. We assign each dialogue either $1$, $2$, or $3$, depending on whether the integrative potential is \textit{low}, \textit{medium}, or \textit{high}. The maximum joint points possible in these cases are $36$, $39$, and $42$ respectively. We find that the participants are able to discover this integrativeness, thereby achieving significantly more joint points as the potential increases ($r=0.425$, $p<0.001$).

\noindent\textbf{Use of negotiation strategies}:
Overall, we find that greater use of Prosocial strategies shows a general pattern to predict higher ratings for both subjective measures of satisfaction and likeness, for self as well as the partner. Engaging in small talk shows significant positive correlations ($ps<0.01$), confirming our hypothesis from prior work that it relates to healthier relationships among the negotiators. Similar effects are visible for \textbf{No-Need} ($ps<0.05$), where the participant decides to let go one of their low-priority items. Since this directly benefits the opponent, it is likely to improve the participant's perception. On the other hand, Proself strategies show a general pattern to predict lower satisfaction and likeness ratings for both self and the partner. We observe significant negative correlation for both \textbf{Other-Need} and \textbf{Vouch-Fair} ($ps<0.01$). Further, we find that these competitive strategies are also associated with lower points scored by the participant and the opponent, and hence, the joint points ($ps<0.01$). These correlations are not influenced by the integrative potential in the scenario, as when the integrated potential is controlled for, the effects generally remain unchanged and demonstrate the same patterns. 

We further observe that the dialogue behavior of a negotiator significantly relates to the behavior of their opponent, where both tend to use similar negotiation strategies ($ps<0.01$). Our findings show that Prosocial strategies are more likely to be associated with Prosocial behavior in the opponents and achieve more favorable outcomes in our negotiation scenario as compared to Proself. These results suggest that an automated negotiator can benefit by employing different strategies based on Prosocial or Proself behaviors of the opponent, for instance, by matching Prosocial behaviors but not Proself. The first step in this direction is to recognize them in a given utterance, which is our focus in the next section.
\section{Strategy Prediction}
\label{sec:modelling}

For building an automated dialogue system that incorporates the negotiation strategies discussed above, an important first step is to build computational models that recognize their usage in the observed utterances. Hence, we explore the task of strategy prediction, given an utterance and its previous dialogue context.

\subsection{Methodology}
\begin{figure}[t!]
\centering
 \includegraphics[width=\linewidth]{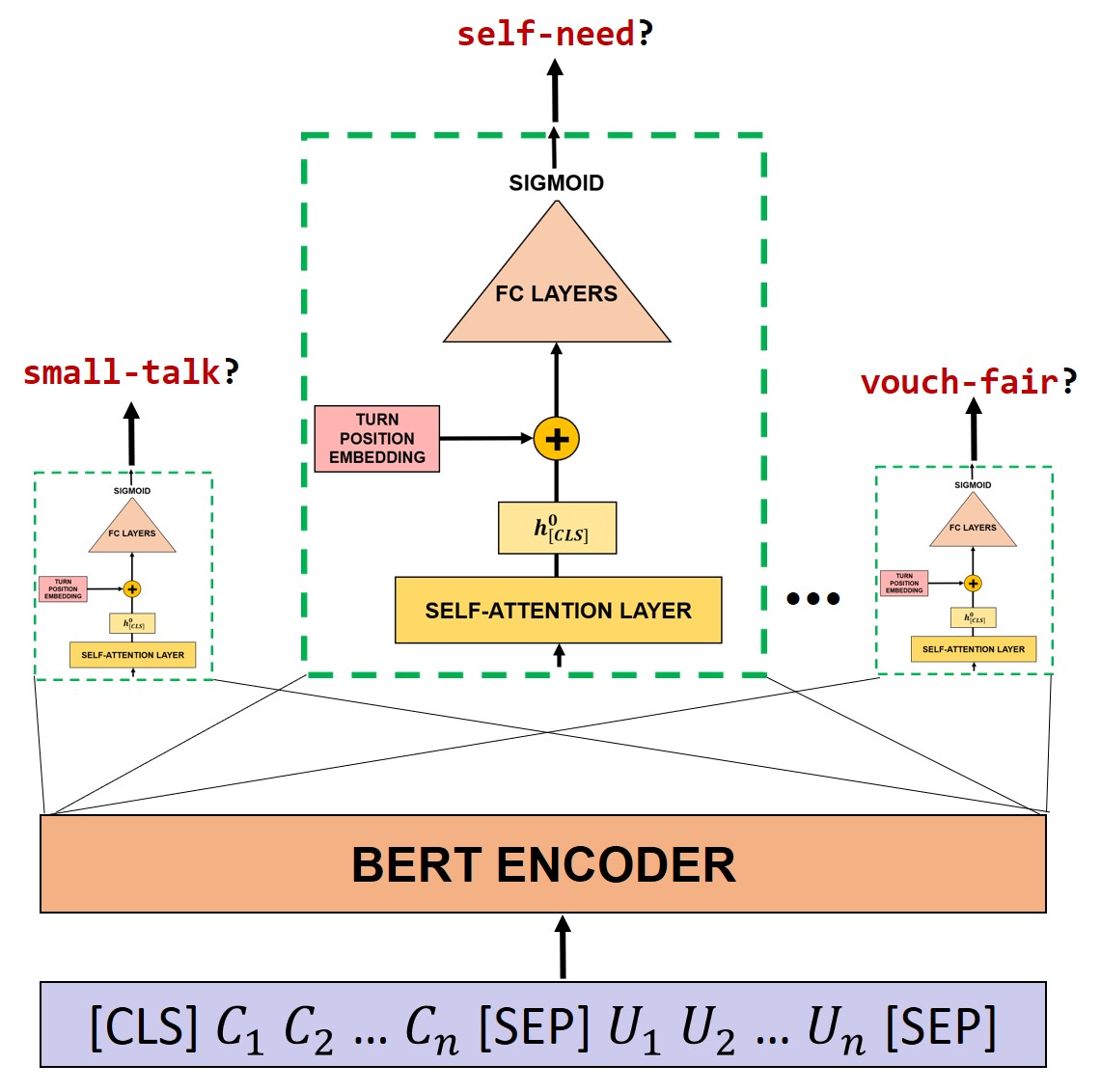}
\caption{Architecture for multi-task strategy prediction. \textbf{+} represents element-wise summation.}
\label{fig:architecture}
\end{figure}
Pre-trained models have proved to be useful on a number of supervised tasks with limited in-domain datasets. Inspired by this success, we use BERT-base~\cite{devlin2019bert} as the core encoding module. A natural way to use pre-trained models for our task is to fine-tune the model for every label independently in a binary classification setup, where the positive class represents the presence of a strategy, and the negative represents its absence. However, most of the utterances in the CaSiNo-Ann dataset are \textbf{Non-strategic}, resulting in a high imbalance where most of the data points belong to the negative class. As we later show, directly fine-tuning the BERT model fails to recognize the strategies for which the data is most skewed. 

We instead propose a multi-task learning framework to allow parameter sharing between the different prediction tasks. Our architecture involves a common BERT-base encoder shared with all the tasks but uses task-specific self-attention to allow the model to focus on the most relevant parts of the input for each task separately. Consequently, this also enables interpretability by allowing us to visualize which parts of an utterance are attended for any given strategy. Our input consists of a finite size context window, which loses the turn index for a specific utterance. Hence, we also capture the turn position for each utterance using sinusoidal positional embeddings~\cite{vaswani2017attention}. We present the complete architecture in Figure \ref{fig:architecture}.

\textbf{In-Domain Pre-Training (IDPT)}: CaSiNo-Ann is nearly $40\%$ of the entire CaSiNo dataset. To incorporate the unannotated dialogues, we employ In-Domain Pre-training of the BERT encoder~\cite{sun2019fine}. For this purpose, we consider each unannotated dialogue as a separate sequence and fine-tune the BERT-base architecture on the Masked Language Modelling (MLM) objective~\cite{devlin2019bert}. This allows us to use the complete CaSiNo dataset in a principled way.

\begin{table*}[ht]
\centering
\scalebox{0.7}{
\begin{tabular}{c|c|c|c|c|c|c|c|cc}
\hline
\textbf{Model} & \multicolumn{1}{c|}{\textbf{Small-Talk}} & \multicolumn{1}{c|}{\textbf{Self-Need}} & \multicolumn{1}{c|}{\textbf{Other-Need}} &
\multicolumn{1}{c|}{\textbf{No-Need}} &
\multicolumn{1}{c|}{\textbf{Elicit-Pref}} &
\multicolumn{1}{c|}{\textbf{UV-Part}} &
\multicolumn{1}{c|}{\textbf{Vouch-Fair}} &
\multicolumn{2}{c}{\textbf{Overall}} \\
& \textbf{F1} & \textbf{F1}  & \textbf{F1} &  \textbf{F1} & \textbf{F1} & \textbf{F1} & \textbf{F1} & \textbf{F1} & \textbf{Joint-A} \\ \hline
\textbf{Majority} & $0.0$ & $0.0$ & $0.0$ & $0.0$ & $0.0$ & $0.0$ & $0.0$ & $0.0$ & $39.6$ \\
\textbf{LR-BoW} & $64.6$ & $57.2$ & $43.2$ & $17.5$ & $56.5$ & $14.3$ & $50.4$ & $43.4$ & $52.4$\\
\textbf{BERT-FT} & $81.6$ & $72.3$ & $76.7$ & $16.4$ & $80.5$ & $20.4$ & $61.9$ & $58.5$ & $64.0$ \\ \hline
\multicolumn{10}{c}{\textbf{Multi-task training}}\\
\textbf{Freeze} & $81.0$ & $69.1$ & $69.5$ & $14.8$ & $77.6$ & $9.2$ & $66.3$ & $55.4$ & $65.8$ \\
\textbf{No Attn} & $80.7$ & $71.9$ & $76.8$ & $7.5$ & $79.0$ & $23.2$ & $60.6$ & $57.1$ & $67.8$ \\
\textbf{No Feats} & $\mathbf{82.7}$ & $75.1$ & $78.8$ & $37.8$ & $82.4$ & $46.2$ & $66.8$ & $67.1$  & $69.9$ \\
\textbf{Full} & $\mathbf{82.7}$ & $74.4$ & $77.9$ & $36.4$ & $\mathbf{83.2}$ & $44.5$ & $\mathbf{67.9}$ & $66.7$ & $\mathbf{70.2}$ \\  \hline
\textbf{+OS} & $82.0$ & $\mathbf{77.1}$ & $75.6$ & $44.2$ & $81.9$ & $46.4$ & $67.3$ & $67.8$ & $70.1$ \\ 
\textbf{+IDPT} & $82.6$ & $74.0$ & $\mathbf{80.4}$ & $41.2$ & $82.8$ & $40.8$ & $64.0$ & $66.6$ & $69.5$ \\
\textbf{+IDPT+OS} & $82.6$ & $75.2$ & $78.8$ &  $\mathbf{46.2}$ & $81.8$ & $\mathbf{47.3}$ & $66.1$ & $\mathbf{68.3}$ & $\mathbf{70.2}$ \\ \hline

\end{tabular}}
\caption{\label{tab:strategy-prediction}Performance on strategy prediction task for $5$-fold cross validation. F1 score corresponds to the positive class.}
\end{table*}

\begin{figure*}[ht]
\centering
\vspace{-0.5cm}
 \includegraphics[width=0.8\linewidth]{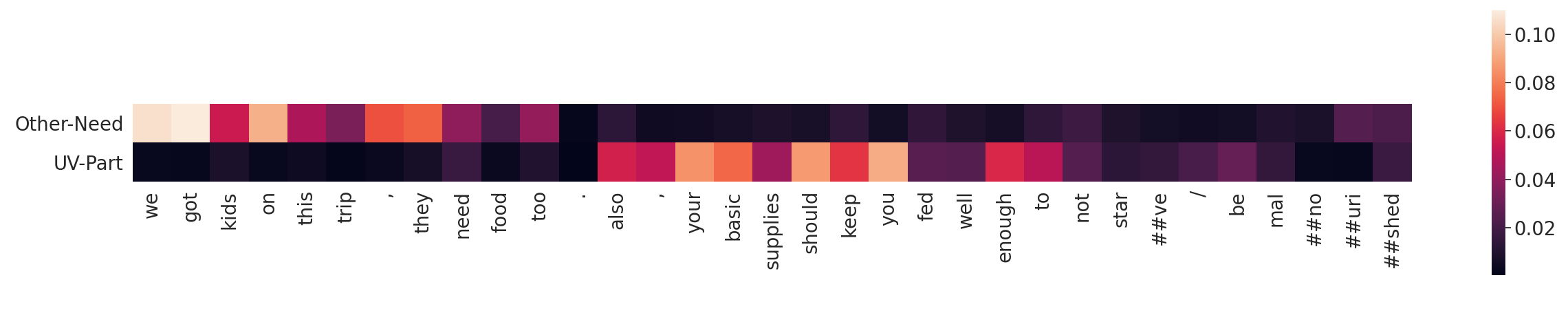}
 \vspace{-0.3cm}
  \includegraphics[width=0.5\linewidth]{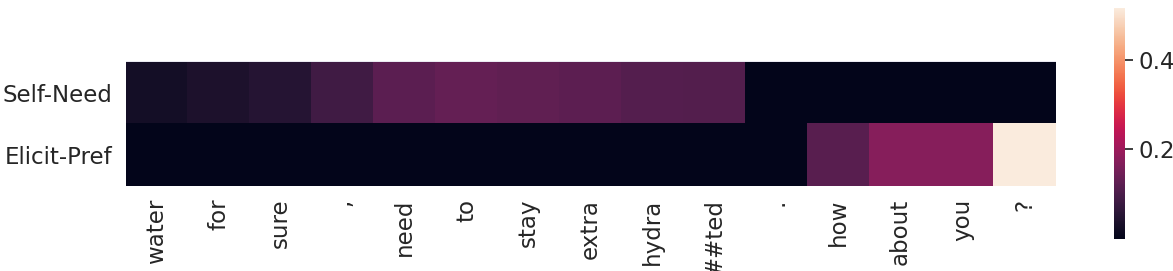}
\caption{Visualizing task-specific self-attention layers for two examples from the test dataset for the first cv fold. The heatmap shows the attention scores for each token in the utterance for corresponding strategy labels.}
\label{fig:attn-viz}
\end{figure*}

\subsection{Experiment Design}

\noindent\textbf{Evaluation Metrics}: We compare our methods for each strategy label on F1-score for positive class (presence of strategy label). To capture the overall performance, we report average F1 across all labels with uniform weights. Inspired by Joint Goal Accuracy from Dialog State Tracking~\cite{kumar2020ma}, we define another overall metric called \textbf{Joint-A}, which measures the percentage of utterances for which the model predicts all the strategies correctly.

\noindent\textbf{Methods}: Fine-tuning the pre-trained models has achieved state-of-the-art results across many supervised tasks. Hence, our primary baseline is \textbf{BERT-FT}, which fine-tunes the BERT-base architecture for binary classification of each strategy label separately. We consider a \textbf{Majority} baseline, where the model directly outputs the majority class in the training data. We also implement a Logistic Regression model for each label separately based on a bag-of-words feature representation of the input utterance. We refer to this model as \textbf{LR-BoW}. We refer to our complete architecture presented in Figure \ref{fig:architecture} as \textbf{Full}, and consider its ablations by freezing the BERT layer (\textbf{Freeze}), removing task-specific self-attention (\textbf{No Attn}), or removing the turn position embeddings (\textbf{No Feats}). We also implement a simple over-sampling strategy where every utterance with at least one strategy is considered twice while training (referred to as \textbf{OS}). For \textbf{IDPT}, we fine-tune BERT for $20$ epochs using a masking probability of $0.3$. We also tried a lower masking probability of $0.15$, however, in that case, the model is unable to learn anything useful on our relatively small dataset.

\noindent\textbf{Training Details}: Our context window considers past $3$ utterances and concatenates them using an \textit{EOS} token. The embedding dimension is $768$ for the encoder and the task-specific self-attention layers, each having only one attention head. We use the turn position embeddings of $32$ dimensions. We train the models with Adam optimizer with a learning rate of $5e^{-05}$ and weight decay of $0.01$. We use ReLU activation for feed-forward layers, and a dropout of $0.1$ to prevent overfitting. The models were trained for a maximum of $720$ iterations with a batch size of $64$ ($\sim13$ epochs). We checkpoint and evaluate the model after every $72$ iterations and the best performing checkpoint on a held-out $5\%$ validation set is used for evaluation. We provide further training details including specifics of the architecture design, computing infrastructure, and hyper-parameter tuning in Appendix F.

\noindent\textbf{Results}: Table \ref{tab:strategy-prediction} summarizes the results on $5$-fold cross-validation. \textbf{Majority} baseline fails to recognize any of the strategies due to the data being skewed towards the negative class. It still achieves $39.4\%$ \textbf{Joint-A}, indicating that these many utterances have none of the seven strategies present. Incorporating the bag-of-words features, \textbf{LR-BoW} performs much better than \textbf{Majority}. \textbf{BERT-FT} highly improves the performance on all strategies except \textbf{No-Need} and \textbf{UV-Part}, for which the dataset is the most skewed. However, our \textbf{Full} multi-tasking framework is able to tackle the imbalance in these strategies through parameter sharing between all tasks. It achieves $36.4\%$ F1 for \textbf{No-Need} and $44.5\%$ F1 for \textbf{UV-Part}, indicating more than $100\%$ improvements in both the cases. The model also improves F1 scores for all other metrics, but the improvement is not that substantial. Relatively lower scores for \textbf{Freeze} and \textbf{No Attn} suggest that both fine-tuning and task-specific attention layers are essential for the performance. Turn position embeddings, however, only help for a few strategies, indicating the diverse usage of strategies in CaSiNo-Ann. Overall, we find that using over-sampling and in-domain pre-training further helps the performance, especially for \textbf{No-Need} and \textbf{UV-Part}. Although there is no clear winner among \textbf{OS} and \textbf{IDPT}, our final model, \textbf{Full+IDPT+OS}, that combines both these strategies performs the best for us, achieving an overall F1 score of $68.3\%$ and $70.2\%$ Joint Accuracy. 

\noindent \textbf{Attention Visualization}:
To understand if the model learns meaningful representations, we visualize the task-specific self-attention layers of the trained \textbf{Full+IDPT+OS} model. We consider two instances in Figure \ref{fig:attn-viz}. For meaningful comparisons, the instances were picked randomly from the pool of all utterances that contain two strategies. As evident, the model is able to focus on the most relevant parts for each strategy label. For instance, in case of \textbf{Other-Need}, the scores are higher where the participant talks about their kids needing more food. The token \textit{we} gets the most attention, which is commonly used by the participants when referring to group needs. We see similar trends in the second case as well. Remarkably, this suggests that although our annotations are at an utterance level, it might be possible to automatically retrieve the most relevant phrases for any given strategy $-$ this requires further investigation which we aim to explore in the future.

\section{Related Work}

Historically, negotiations have been widely studied across multiple disciplines, in game theory~\cite{nash1950bargaining}, understanding human behaviour~\cite{adair2001negotiation}, and building automatic negotiation agents~\cite{beam1997automated, baarslag2016survey}. Most efforts focused on agent-agent interactions~\cite{williams2012iamhaggler, Genius, cao2018emergent}, although there is an increasing interest in human-agent negotiations~\cite{mell2017grumpy} as well.
\citet{devault2015toward} used a multi-issue bargaining design similar to ours. However, they focus on face-to-face negotiations, including speech and virtual embodied systems, which can be interesting future extensions to our current focus in chat-based dialogue systems. Other datasets looked at negotiation dialogues such as game settings~\cite{asher2016discourse,lewis2017deal}, and buyer-seller negotiations~\cite{he2018decoupling}. These datasets have fueled a number of efforts on developing negotiation systems~\cite{cheng2019evaluating, parvaneh2019show} and building a negotiation coach~\cite{zhou2019dynamic}. Our focus is on campsite negotiations, targeting a realistic and a closed-domain environment.

Several other related efforts have explored problems between task-oriented and open-domain scenarios, such as persuasion for a charity~\cite{wang2019persuasion}, anti-scam~\cite{li2020end}, collecting cards in a maze~\cite{potts2012goal}, and searching for a mutual friend~\cite{he2017learning}. Instead, we focus on rich personal negotiations, which differ from these tasks in their ultimate goal and downstream applications.

\section{Conclusions and Future Work}
\label{sec:conclusion}
We described the design and development of the CaSiNo dataset and the associated annotations. Our design is based on a relatable campsite scenario that promotes constrained, yet linguistically rich and personal conversations. We next plan to explore two main projects: first, extending the analysis to demographic and personality traits in the data, and second, using our insights towards the development of practical automated negotiation systems that engage in free-form dialogue and portray well-studied strategies from the prior negotiation literature. Our work fuels other tasks to advance the research in human-machine negotiations, such as predicting satisfaction and opponent perception from dialog behaviors, and building a feedback mechanism for skills training by identifying the use of pro-social versus pro-self strategies.

Finally, we note that there are many interesting extensions to our task design that make the scenario more complicated, but useful in specific realistic settings. For instance, incorporating more than two negotiating parties, and considering other modalities like facial expressions or embodied agents. In some realistic settings, the individual preferences may change during the negotiation and our setup assumes a fixed set of preferences throughout. Further, in complex settings, it may be possible to break down an individual item and claim sub-parts, such as negotiating for who gets an orange, but one party ends up taking the husk and the other takes the pulp for their own purposes. This is again not considered in our work and opens up exciting avenues for future work.

\section{Broader Impact and Ethical Considerations}
\label{sec:ethics}

\subsection{Data Collection}

Our study was approved by our Institutional Review Board (IRB). Each participant signed an Informed Consent document at the beginning of the study which covered the purpose of the study, warned about potential discomfort, and noted the collection of data and its later use. Further, the participants were informed that they can withdraw at any time. They were also instructed to not use any offensive or discriminative language. The compensation was determined in accordance with the fairness rules defined by our IRB approval process. Additionally, we release the anonymized version of the data for future work by the research community. All personally identifiable information such as MTurk Ids or HIT Ids was removed before releasing the data. Lastly, any mention of the demographics or the psychological personality of the participants is based on self-identified information in our pre-survey and standard procedures of collecting personality metrics in the literature.

\subsection{Automatic Negotiation Systems}

Students entering the modern workforce must have a number of interpersonal skills that are crucial across a wide range of jobs. One of the key interpersonal skills needed to address conflicts and work well with others is the ability to negotiate. Unfortunately, research shows that the average human is bad at negotiating. This can adversely impact work opportunities~\cite{babcock2009women}, legal settlements~\cite{eisenberg2009settlement}, and cross-cultural border peace~\cite{wunderle2007negotiate}. The typical way to teach negotiation skills to students is by in-class simulations, which are expensive. Automated systems can dramatically reduce the costs of, and increase access to, negotiation training. Systems developed on CaSiNo would be useful in this context. Further, the techniques developed find use-cases for advancing conversational AI and imparting the negotiation skills to existing AI assistants, making them more aware of our preferences and requirements. One such prototype is Google Duplex~\cite{leviathan2018google}, where the AI system engages in a simple form of negotiation to book a haircut appointment over the phone.

How humans negotiate has been actively studied for decades in Economics, Psychology, and Affective Computing~\cite{carnevale1992negotiation}. With this huge progress in our understanding of human-human negotiations, ethics has also been a well-studied topic in the literature~\cite{lewicki2016essentials}. Primary concerns include the acts of emotion manipulation, deception, bias, and misrepresentation. Naturally, these ethical concerns may creep into the automated systems, trained on a human-human negotiation dataset.

To mitigate these ethical impacts, we recommend that standard guidelines for deploying conversational AI assistants should be followed. It is essential to maintain transparency about the identity of the system. Ethical principles must be in place before the deployment of such systems with a regular update cycle. Our camping scenario is quite relatable to anyone who negotiates with the system, hence, it is important to be upfront about the potential behaviors of the deployed system. We recommend continuous monitoring by keeping humans in the loop, ensuring that the system is neither offensive nor discriminative. Further, it should be made easy for the users negotiating with the system to directly contact the team behind the deployment. Finally, any data which is collected during the deployment phase should be informed to the users and its future purpose should be properly laid out.

\section*{Acknowledgments}
We would like to thank Shivam Lakhotia, along with colleagues at the Institute for Creative Technologies and Information Sciences Institute for their comments and helpful discussions. We further thank Mike Lewis, He He, Weiyan Shi, and Zhou Yu for their guidance. We also thank the anonymous reviewers for their valuable time and feedback. Our research was sponsored by the Army Research Office and was accomplished under Cooperative Agreement Number W911NF-20-2-0053. The views and conclusions contained in this document are those of the authors and should not be interpreted as representing the official policies, either expressed or implied, of the Army Research Office or the U.S. Government. The U.S. Government is authorized to reproduce and distribute reprints for Government purposes notwithstanding any copyright notation herein.


\bibliography{anthology,custom}
\bibliographystyle{acl_natbib}

\newpage

\appendix

\section{Pre-Survey}
\label{sec:pre-survey-appendix}
After an internal pilot with $9$ participants, the entire CaSiNo dataset was collected on Amazon Mechanical Turk over a period of a month. In total, $846$ subjects took part in our data collection study. The statistics presented in this section are based on self-identified demographical attributes and standard ways of collecting personality traits from the literature. We had a highly diverse participant pool, representing different age groups, gender, ethnic backgrounds and education levels. The mean \textit{Age} among our participants is $36.97$ with a standard deviation of $10.81$. One participant was removed from this computation since the age entered was $3$, which we believed to be in error. Among the participants, $472$ identified themselves as \textit{Female}, $372$ were \textit{Male}, and \textit{2} belonged to \textit{Other} category. While most of the participants were \textit{White American} ($625$ in count), our study also involved a mix of \textit{Asian American}, \textit{Black or African American}, \textit{Hispanic or Latino}, and \textit{Multi-Racial} groups, among others. Most common \textit{highest level of education} was found to be a $4$-year Bachelor degree ($346$ participants), although the complete pool represents a mixture of Master and PhD degree holders, $2$-year and $4$-year college graduates without degrees, and high school graduates, among others.

For the personality traits, $364$ participants were classified as Proself, $463$ as Prosocial, and $19$ were unclassified based on their Social Value Orientation\footnote{\url{https://static1.squarespace.com/static/523f28fce4b0f99c83f055f2/t/56c794cdf8baf3ae17cf188c/1455920333224/Triple+Dominance+Measure+of+SVO.pdf}}. The mean scores for the Big-$5$
 personality traits were found to be as follows: Agreebleness: $5.27$, Conscientiousness: $5.6$, Emotional Stability:  $4.91$, Extraversion: $3.69$, Openness to Experiences: $5.04$. We use the Ten-Item Personality Inventory (TIPI)\footnote{\url{https://gosling.psy.utexas.edu/scales-weve-developed/ten-item-personality-measure-tipi/ten-item-personality-inventory-tipi/}} to compute these attributes, where each of them takes a value between $1$ and $7$.

\section{Preparation Phase}
\label{sec:prep-phase-appendix}

We present the scenario description seen by the participants in Table \ref{tab:scenario}. Several arguments that the participants come up with are presented in Table \ref{tab:sample-reasons}.

\begin{table}[ht]
\centering
\scalebox{0.9}{
\begin{tabular}{|c|}
\hline
\pbox{8cm}{\vspace{0.1cm}\textit{Imagine that you are on a camping trip! Woohoo!}\\\textit{Apart from some basic amount of supplies which are provided to everyone, you can collect some additional food packages, water bottles and firewood, to make your camping trip even better. Since these are limited in quantity, you will have to split these additional packages with your campsite neighbor!}\\\textit{Each of these items will be of either High, Medium or Low priority for you. Each of them only has an available quantity of 3. You will negotiate with another MTurker by chatting in English, using reasons from your personal experiences to justify why you need additional packages apart from the basic supplies. Try hard to get as many items as you can!}\vspace{0.1cm}}
\\ \hline
\end{tabular}}
\caption{The camping scenario description as seen by the participants in our data collection.}
\label{tab:scenario}
\end{table}

\begin{table*}[ht]
\centering
\scalebox{0.8}{
\begin{tabular}{l|P{5cm}|P{5cm}|P{5cm}}
\hline
\textbf{Category} & \multicolumn{3}{c}{\textbf{Item type}} \\ \hline
& \textbf{Food} & \textbf{Water} & \textbf{Firewood} \\ \hline

\textbf{Personal Care} & because I'm normally eat more because of my big size & I have to take a lot of medicine so hydration is very important &  I have arthritis and being sure I am warm is important for my comfort. \\ \hline
\textbf{Recreational} & Need many snacks throughout the day for energy to hike & I am a very active camper. I like to hike when I camp and I once ran out of water during a strenuous hike. &  I like having campfires so I need all the firewood.\\ \hline
\textbf{Group Needs} & I have two teenage boys who require a lot of food, especially when expending so much energy with all the activities of camping. &  I need more water because I have more people to keep hydrated and do not have enough. & I need more firewood due to having several people join on the trip and needing a bigger fire overall. \\ \hline
\textbf{Emergency} & Some could have been damaged during the trip. I would need more. & our car overheated we had to use the water &  It may get cold and firewood can be hard to come by at certain campsites. 
 \\ \hline
 
\end{tabular}}
\caption{Example arguments that the participants come up for their individual requirements during the preparation phase. The categories defined are not exhaustive.}
\label{tab:sample-reasons}
\end{table*}

\section{Data Post-processing steps}
\label{sec:post-processing-appendix}

We list the data post-processing and filtering steps below:
\begin{enumerate}
    \item \textbf{Removal of incomplete dialogues}: During the data collection, many negotiation sessions could not be completed due to one of the participants' disconnecting in the middle. Any dialogue for which we had missing data, including pre-survey and post-survey responses for both the participants, was removed from the final dataset.
    \item \textbf{Removal of bad quality dialogues}: We also removed dialogues where we observed a lack of effort or an irrelevant dialogue between the participants. We removed dialogues where the participants used very short utterances or failed to answer the dummy questions about their own preferences correctly, suggesting a lack of effort. Further, we removed the instances where the participants talked about the MTurk task itself, rather than the negotiation. These cases were identified based on a list of keywords: $\{$`mturk', `amt', `turns', `messages', `amazon', `10'$\}$. In a few cases, it was possible to retain the complete dialogue structure by just removing a few utterances. Hence, in these cases, we only removed the irrelevant utterances, while retaining the rest of the dialogue and the associated metadata.
    \item \textbf{Tackling inappropriate language use}: Rarely, some participants also used inappropriate language in their utterances. These dialogues were identified using the lexicon of English swear words on Wikipedia\footnote{\url{https://en.wiktionary.org/wiki/Category:English_swear_words}}. All these dialogues were also removed from the final dataset.
    
\end{enumerate}

\section{Participant Feedback}
\label{sec:feedback-appendix}

Role-playing has been a key technique to teach negotiation skills in classroom settings. One of the key application areas for automated negotiation systems is to augment such exercises by allowing the human participants to negotiate with an AI and practice their social skills. To maximize the utility of the system developed using our dataset, we choose the camping scenario, which we expected to be easily relatable for our participants and also for any individual who negotiates with a system developed on our dataset. This is essential to ensure that the collected dialogues are engaging, interesting, and capture the rich personal context of the individuals, albeit in a closed-domain setting. 
One way to judge whether the participants are able to relate to the scenario is via their feedback after the study. With this in mind, we used a feedback column in the Post-survey and asked several questions to the participants throughout the data collection process. These questions included: 1) How was your overall experience? 2) Were you able to see yourself in the `role' and follow best practices?, 3) Could you relate to camping?, and 4) How helpful was the preparation phase?

Based on manual inspection, we observed an overall positive feedback for all the above questions. Most of the participants were able to easily relate to camping. They frequently pointed out that the experience was `fun', `interesting', and `nice'. Many saw this as an opportunity to talk to someone during these tough times of the pandemic. Several cherry-picked feedback responses which indicate that the participants enjoyed the task as a whole and were in fact able to connect well and engage in the negotiation, have been provided in Table \ref{tab:feedback}.

\begin{table*}[ht]
\centering
\scalebox{0.8}{
\begin{tabular}{p{15cm}}
\hline
I could do this all day \\ \hline
I am camping right now!\\ \hline
My partner had better reasons for needing the firewood \\ \hline
I enjoyed talking about camping, I haven't been in a while. It reminded me of all of the things that I used to do.\\ \hline
The best thing I did was ask him what his preferences were. He had no interest in firewood which was my highest priority. \\ \hline

\end{tabular}}
\caption{A few positive feedback responses which we obtained from the participants during the collection of the CaSiNo dataset.} 
\label{tab:feedback}
\end{table*}

\section{Correlational Analysis}
\label{sec:corr-analysis-appendix}
The analysis discussed in the paper is presented in Tables \ref{tab:corr-bw-outcomes}, \ref{tab:t-test-integrative}, \ref{tab:corr-annos}, and  \ref{tab:corr-annos-among-itself}.


\begin{table*}[ht]
\centering
\scalebox{0.7}{
\begin{tabular}{l|lll}
\hline
& \textbf{Points-Scored} & \textbf{Satisfaction} & \textbf{Opp-Likeness} \\ \hline
\textbf{Points-Scored} & $1$ & $.376$** & $.276$** \\
\textbf{Satisfaction} & $.376$** & $1$ & $.702$** \\
\textbf{Opp-Likeness} &$.276$**  & $.702$** & $1$ \\ \hline
\textbf{P.Points-Scored} &$-.092$**    & $.105$**    & $.132$**        \\
\textbf{P.Satisfaction} &$.105$**   & $.180$**  & $.244$**   \\
\textbf{P.Opp-Likeness} & $.132$** & $.244$**& $.344$**     \\ \hline
\end{tabular}}
\caption{\label{tab:corr-bw-outcomes} Pearson Correlation Coefficients (r) between the outcome variables. Variables with \textbf{P.} prefix denote the corresponding attributes of the negotiation partner of an individual. These correlations have been computed on the entire CaSiNo dataset. * denotes significance with $p<0.05$ ($2$-tailed). ** denotes significance with $p<0.01$ ($2$-tailed).}
\end{table*}

\begin{table*}[ht]
\centering
\scalebox{0.8}{
\begin{tabular}{l|l}
\hline
& \multicolumn{1}{c}{\textbf{Joint Points}} \\ \hline
\textbf{Integrative potential} & $.425$*** \\ \hline
\end{tabular}}
\caption{\label{tab:t-test-integrative} Pearson Correlation Coefficient (r) between integrative potential and the joint negotiation performance. *** denotes significance with $p<0.001$.}
\end{table*}

\begin{table*}[ht]
\centering
\scalebox{0.7}{
\begin{tabular}{l|lllllll}
\hline
& \textbf{Joint Points} & \textbf{Points-Scored} & \textbf{Satisfaction} & \textbf{Opp-Likeness} & \textbf{P.Points-Scored} & \textbf{P.Satisfaction} & \textbf{P.Opp-Likeness} \\ \hline
& \multicolumn{7}{c}{\textbf{Prosocial Generic}} \\
\textbf{Small-Talk} & $-.022$ & $-.002$        & $.086$*       & $.115$**           & $-.025$                & $.068$                & $.127$**  \\ \hline
& \multicolumn{7}{c}{\textbf{Prosocial About Preferences}} \\
\textbf{No-Need} & $-.003$ & $-.066$        & $.035$        & $.023$             & $.063$                 & $.083$*               & $.089$*  \\
\textbf{Elicit-Pref} & $.053$ & $.055$         & $.058$        & $.015$             & $.010$                 & $.022$                & $.055$ \\ \hline
& \multicolumn{7}{c}{\textbf{Proself Generic}} \\
\textbf{UV-Part} & $-.037$ & $.008$         & $-.051$       & $-.112$**          & $-.054$                & $-.131$**             & $-.151$** \\
\textbf{Vouch-Fairness} & $-.140$** & $-.084$*       & $-.159$**     & $-.196$**          & $-.090$*               & $-.185$**             & $-.180$**  \\ \hline
& \multicolumn{7}{c}{\textbf{Proself About Preferences}} \\
\textbf{Self-Need} & $-.003$ & $.022$         & $-.061$       & $-.065$            & $-.026$                & $-.091$*              & $-.086$*  \\
\textbf{Other-Need} & $-.176$** & $-.045$        & $-.101$**     & $-.118$**          & $-.174$**              & $-.160$**             & $-.113$**  \\ \hline
\end{tabular}}
\caption{\label{tab:corr-annos} Pearson Correlation Coefficients (r) for strategy annotation counts with the outcome variables. Variables with \textbf{P.} prefix denote the corresponding attributes of the negotiation partner of an individual. These correlations have been computed on the annotated subset of the CaSiNo dataset. * denotes significance with $p<0.05$ ($2$-tailed). ** denotes significance with $p<0.01$ ($2$-tailed).}
\end{table*}

\begin{table*}[ht]
\centering
\scalebox{0.8}{
\begin{tabular}{l|lllllll}
\hline
& \textbf{P.Small-Talk} & \textbf{P.Self-Need} & \textbf{P.Other-Need} & \textbf{P.No-Need} & \textbf{P.Elicit-Pref} & \textbf{P.UV-Part} & \textbf{P.Vouch-Fair} \\ \hline
\textbf{Small-Talk} & $.769$**   & $-.033$       & $.021$         & $.063$       & $-.059$                & $-.012$                     & $-.180$**  \\
\textbf{Self-Need} & $-.033$    & $.355$**      & $.103$**       & $.115$**     & $-.007$                & $.235$**                    & $-.088$*  \\
\textbf{Other-Need} & $.021$     & $.103$**      & $.339$**       & $.002$       & $-.067$                & $.159$**                    & $-.015$  \\
\textbf{No-Need} & $.063$     & $.115$**      & $.002$         & $.258$**     & $.097$**               & $.064$                      & $-.116$**   \\
\textbf{Elicit-Pref} & $-.059$    & $-.007$       & $-.067$        & $.097$**     & $.168$**               & $-.097$**                   & $-.102$**  \\
\textbf{UV-Part} & $-.012$    & $.235$**      & $.159$**       & $.064$       & $-.097$**              & $.268$**                    & $.064$   \\
\textbf{Vouch-Fair} & $-.180$**  & $-.088$*      & $-.015$        & $-.116$**    & $-.102$**              & $.064$                      & $.287$**   \\ \hline
\end{tabular}}
\caption{\label{tab:corr-annos-among-itself} Pearson Correlation Coefficients (r) between strategy annotation counts. Variables with \textbf{P.} prefix denote the corresponding attributes of the negotiation partner of an individual. These correlations have been computed on the annotated subset of the CaSiNo dataset. * denotes significance with $p<0.05$ ($2$-tailed). ** denotes significance with $p<0.01$ ($2$-tailed).}
\end{table*}

\section{Strategy Prediction}
\label{sec:strategy-prediction-appendix}

\subsection{Architecture}
We provide some more details on the strategy prediction multi-task architecture in this section. The self-attention layer is itself represented using the BERT encoder architecture, but with a single transformer layer and just one attention head. After the self-attention layer, we first extract the $768$ dimensional representation for the [CLS] token. This is passed through a feed-forward network, which converts it to $128$ dimensions. The feature embedding is also converted to a $128$ dimensional vector using a feed-forward network. Both the above embeddings are then combined using an element-wise summation, which further passes through two feedforward layers with hidden dimensions of $64$ and $1$, and a sigmoid layer to finally output the probability for each annotation strategy.

\subsection{Computing Infrastructure}
All experiments were performed on a single Nvidia Tesla V100 GPU. The training takes two hours to complete for a single model on all the cross-validation folds.

\subsection{Training Details}
To search for the best hyperparameters, we use a combination of randomized and manual search for the \textbf{Full} model. For each cross fold, $5\%$ of the training data was kept aside for validation. The metric for choosing the best hyper-parameters is the mean F1 score for the positive class on the validation dataset. The mean is over all the labels and over $5$ cross-validation folds.

We vary the learning rate in $\{3e^{-5}$, $4e^{-5}$, $5e^{-5}\}$, weight decay in $\{0.0$, $0.01$, $0.001\}$ and dropout in $\{0.0$, $0.1$, $0.2$, $0.3\}$. The rest of the hyper-parameters were fixed based on the available computational and space resources. We report the best performing hyper-parameters in the main paper, which were used for all the experiments. We report the performance on the validation set corresponding to the chosen hyper-parameters and the number of trainable parameters in Table \ref{tab:val-perf}.

\begin{table*}[ht]
\centering
\scalebox{0.8}{
\begin{tabular}{c|c|c}
\hline
\textbf{Model} & \textbf{Overall Validation F1} & \textbf{Trainable Parameters} \\ \hline
\textbf{Majority} & $0.0$ & $0$\\
\textbf{LR-BoW} & $49.6$ & $2646.2$ ($27.2$)\\
\textbf{BERT-FT} & $69.9$ & $109,590,529$\\ \hline
& \multicolumn{1}{c}{\textbf{Multi-task training}}& \\
\textbf{Freeze} & $62.3$ & $221,361,031$\\
\textbf{No Attn} & $66.6$ & $110,235,271$\\
\textbf{No Feats} & $77.6$ & $330,840,583$ \\
\textbf{Full} & $78.1$ & $330,844,807$\\  \hline
\textbf{+OS} & $77.9$ & $330,844,807$\\ 
\textbf{+IDPT} & $79.6$ & $330,844,807$\\
\textbf{+IDPT+OS} & $79.6$ & $330,844,807$\\ \hline

\end{tabular}}
\caption{\label{tab:val-perf}Training details for the strategy prediction task. The Overall F1 scores are for the positive class. For \textbf{LR-BoW}, the exact number of features varies slightly based on the CV split. Hence, we report Mean (Std) across the five splits.}
\end{table*}

\section{Screenshots from the data collection interface}
\label{sec:snapshots-appendix}

\begin{figure*}[ht]
\centering
\begin{subfigure}[b]{\textwidth}
 \centering
 \includegraphics[width=\textwidth]{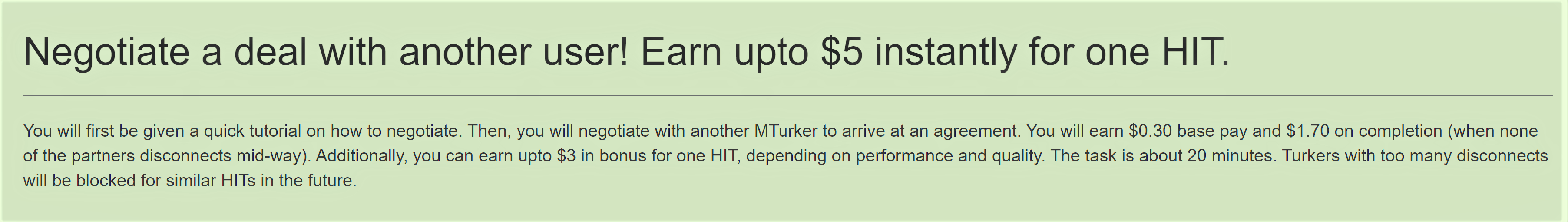}
\end{subfigure}
\caption{Screenshots from the data collection interface: Task Preview. This is a brief task description which the MTurkers see before signing up for our data collection task.}
\label{fig:screenshots1}
\end{figure*}

\begin{figure*}[ht]
\centering
\begin{subfigure}[b]{0.49\textwidth}
 \centering
 \includegraphics[width=\textwidth]{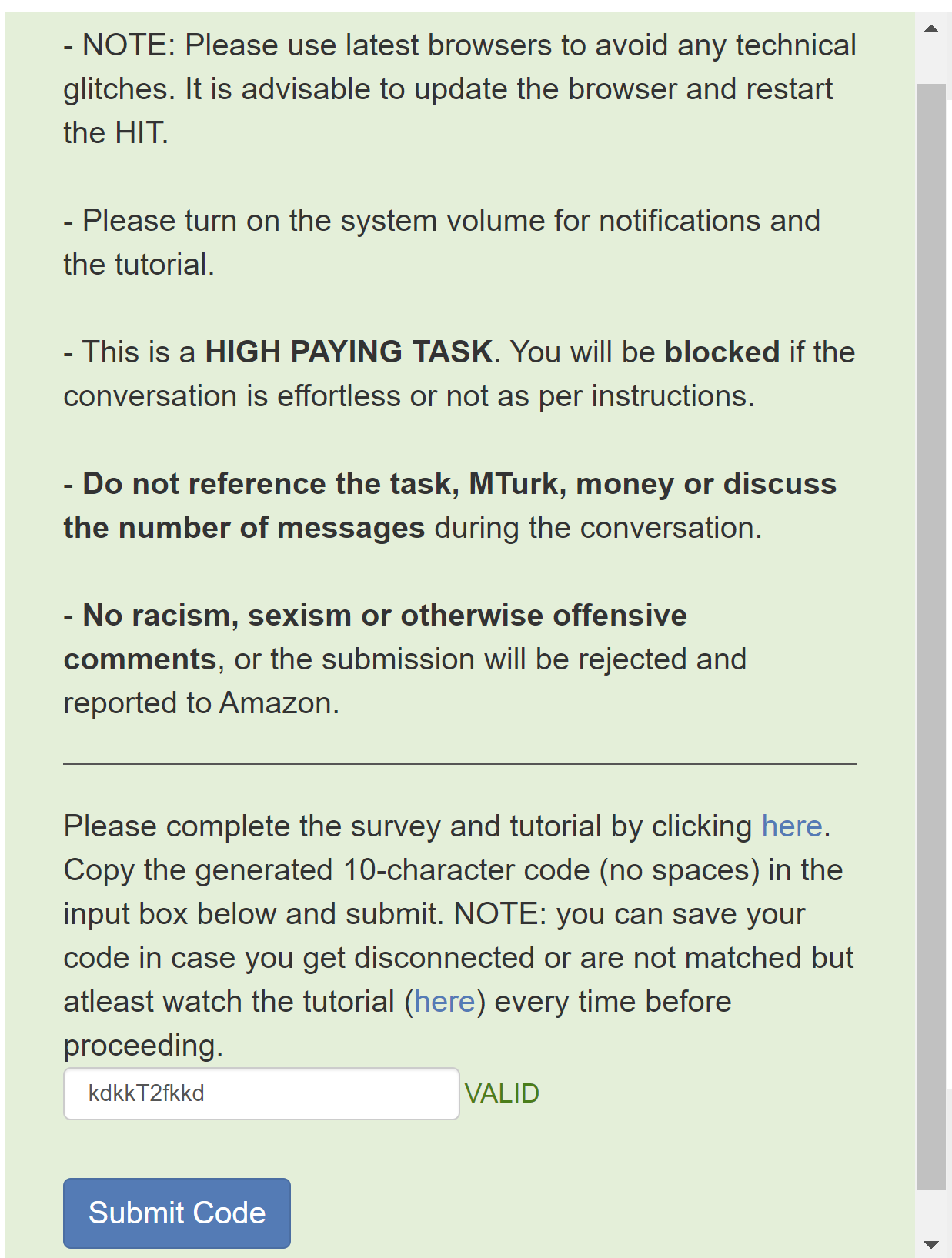}
 \caption{Onboarding Phase 1: The first step takes the participant to Qualtrics which collects the demographics, introduces the camping scenario and gives a tutorial on negotiation best practices.}
\end{subfigure}
\begin{subfigure}[b]{0.49\textwidth}
 \centering
 \includegraphics[width=\textwidth]{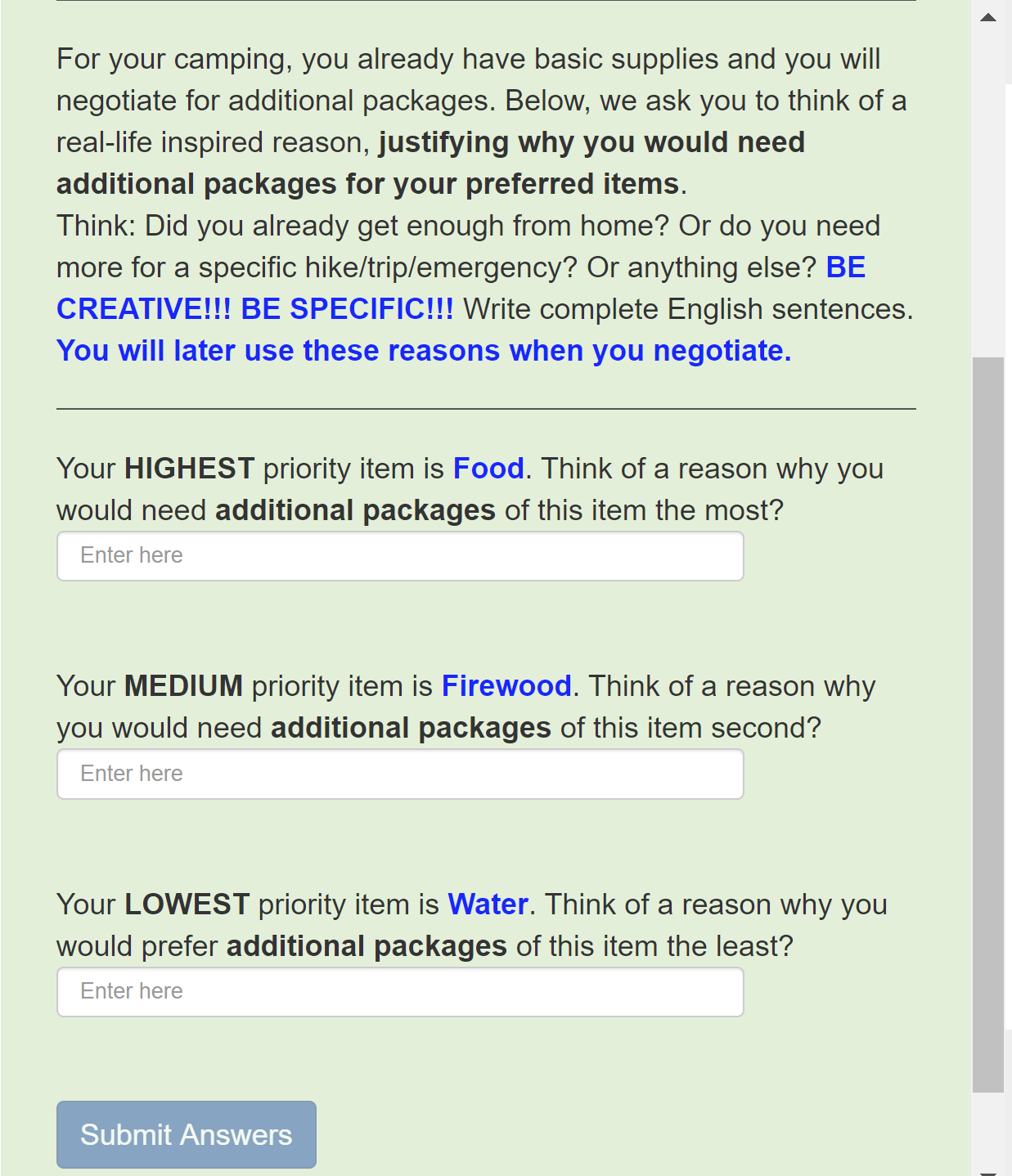}
  \caption{Onboarding Phase 2: In this phase, we explicitly ask the participants to come up with arguments from their past experiences, which justify their preferences. The preference order is randomly assigned by us. This provides a personal context around the negotiation for each participant.}
\end{subfigure}
\caption{Screenshots from the data collection interface: Participant On-boarding.}
\label{fig:screenshots2}
\end{figure*}

\begin{figure*}[ht]
\centering
\begin{subfigure}[b]{\textwidth}
 \centering
 \includegraphics[width=\textwidth]{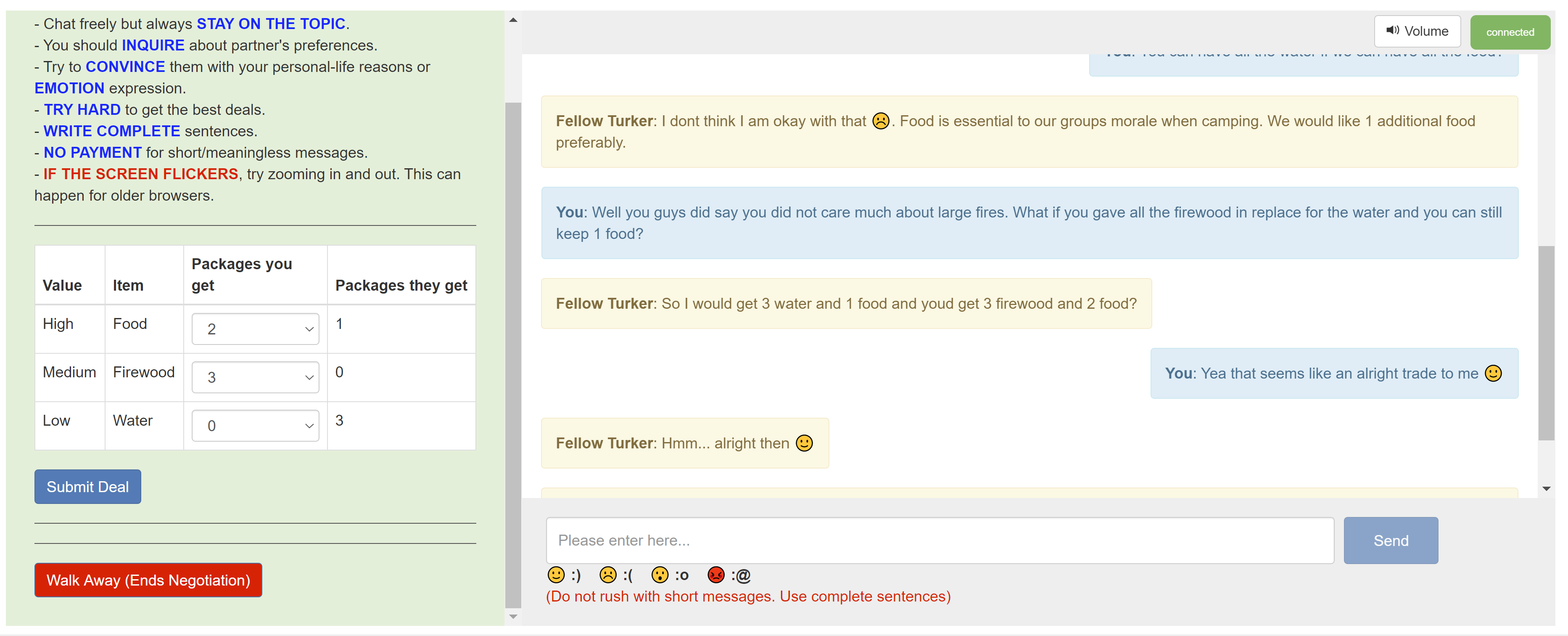}
 \caption{Chat Interface: The right portion allows two participants to negotiate in English using alternating messages. They also have the option to use emoticons. Once they come to an agreement, one of the participant must enter the exact deal on the left.}
\end{subfigure}
\begin{subfigure}[b]{\textwidth}
 \centering
 \includegraphics[width=\textwidth]{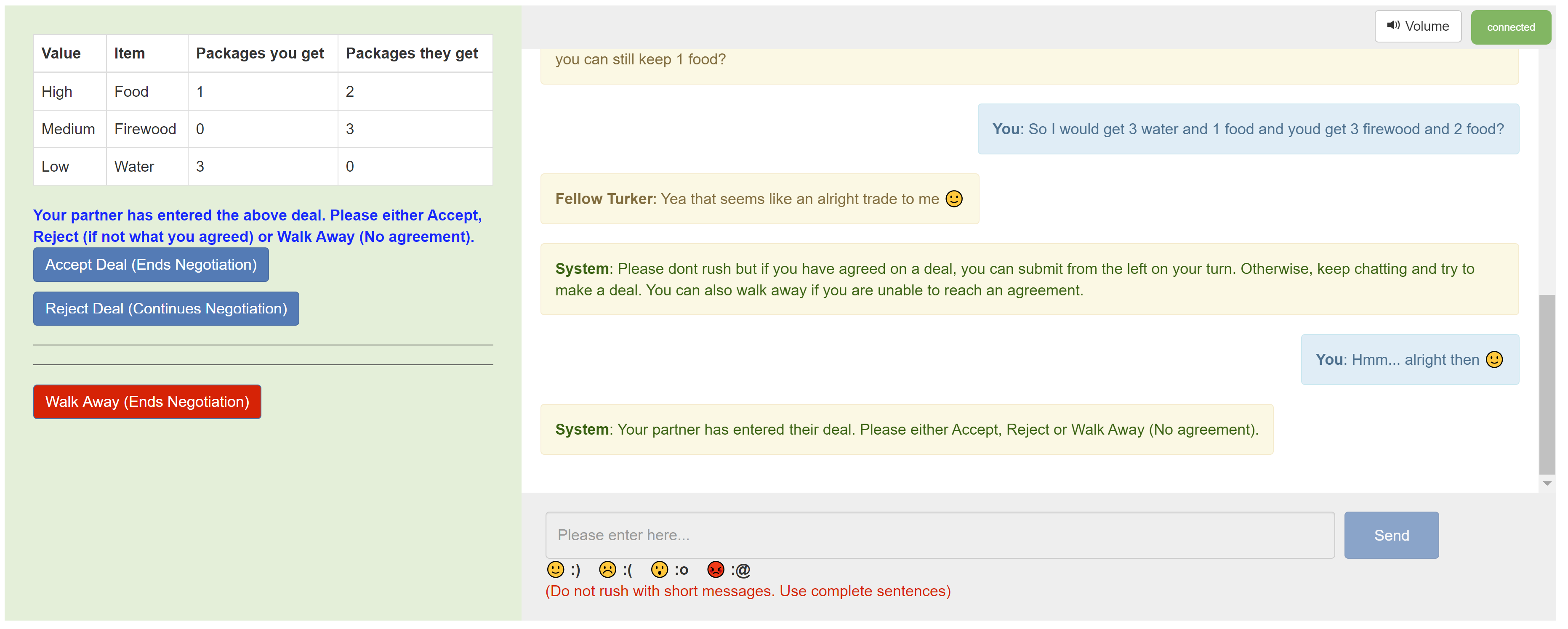}
 \caption{Response to the Deal: When one of the participants enters the deal, the other gets an option to either accept, reject, or walk away from the deal. In the CaSiNO dataset, a participant walks away in $36$ dialogues.}
\end{subfigure}
\caption{Screenshots from the data collection interface: Chat Interface.}
\label{fig:screenshots3}
\end{figure*}

\begin{figure*}[ht]
\centering
\begin{subfigure}[b]{0.6\textwidth}
 \centering
 \includegraphics[width=\textwidth]{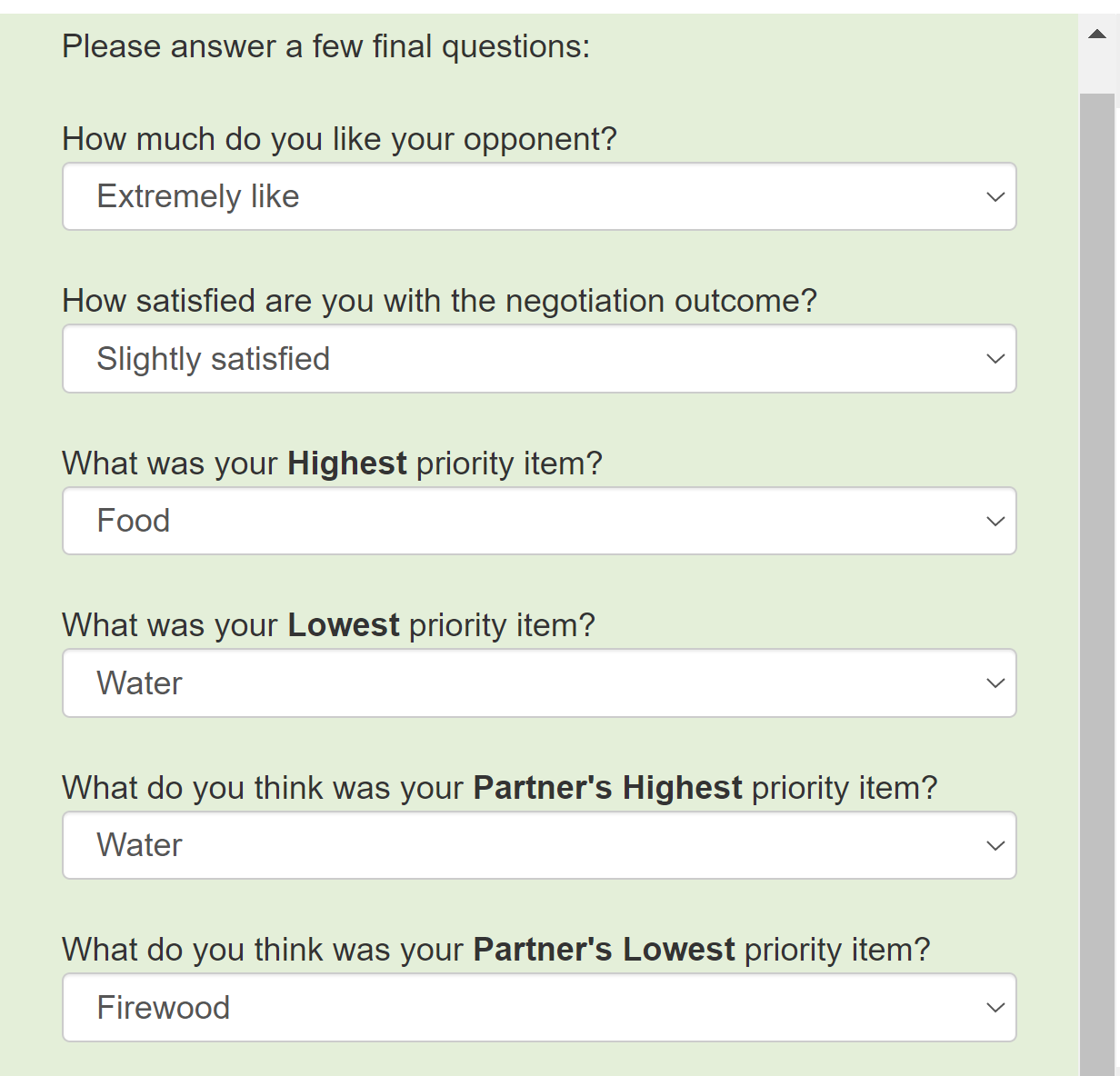}
\end{subfigure}
\caption{Screenshots from the data collection interface: Post-Survey. Once the deal is accepted (or someone walks away), both the participants are asked to fill in the post-survey having the above questions. The figure contains dummy responses.}
\label{fig:screenshots4}
\end{figure*}

To provide more clarity on the data collection procedure, we provide several screenshots from our interface in Figures \ref{fig:screenshots1}, \ref{fig:screenshots2}, \ref{fig:screenshots3}, and \ref{fig:screenshots4}. We design the pre-survey using the Qualtrics platform\footnote{\url{https://www.qualtrics.com/core-xm/survey-software/}}. The rest of the data collection is based on the ParlAI framework~\cite{miller2017parlai}.

\end{document}